\documentclass[runningheads]{llncs}
\usepackage{graphicx}
\usepackage{amsmath,amssymb} 
\usepackage{algorithm, algorithmic}
\usepackage{subfig}
\usepackage{color}
\usepackage{soul}	
\usepackage{float}
\usepackage{multirow,multicol}

\usepackage[hidelinks]{hyperref}
\usepackage{bbm}

\newcommand{\real}{\mathbb{R}}
\newcommand{\se}{\mathrm{SE(3)}}
\newcommand{\shape}{\mathcal{S}}

\def\captionfont{\small}

\def\tablefontsize{scriptsize}

\begin{document}

\pagestyle{headings}
\mainmatter

\title{
Visual-Inertial Object Detection and Mapping 
}

\titlerunning{Visual-Inertial Object Detection and Mapping}

\authorrunning{ X. Fei and S. Soatto}

\author{Xiaohan Fei and Stefano Soatto}

\institute{UCLA Vision Lab\\
	University of California, Los Angeles\\
	\email{ \{feixh,soatto\}@cs.ucla.edu}
}

\maketitle

\begin{abstract}
We present a method to populate an unknown environment with models of previously seen objects, placed in a Euclidean reference frame that is inferred causally and on-line using monocular video along with inertial sensors. The system we implement returns a sparse point cloud for the regions of the scene that are visible but not recognized as a previously seen object, and a detailed object model and its pose in the Euclidean frame otherwise. The system includes bottom-up and top-down components, whereby deep networks trained for detection provide likelihood scores for object hypotheses provided by a nonlinear filter, whose state serves as memory. Additional networks provide likelihood scores for edges, which complements detection networks trained to be invariant to small deformations. We test our algorithm on existing datasets, and also introduce the VISMA dataset, that provides ground truth pose, point-cloud map, and object models, along with time-stamped inertial measurements. 
\end{abstract}

\section{Introduction}
\label{sect-intro}
We aim to detect, recognize, and localize objects in the three-dimensional (3D) scene. We assume that previous views of the object are sufficient to construct a dense model of its shape, in the form of a closed and water-tight surface, and its appearance (a texture map). So, as soon as an object is detected from a monocular image, and localized in the scene, the corresponding region of space can be mapped with the object model, including the portion not visible in the current image (Fig.~\ref{fig-visual} and~\ref{fig-scenenn-visual}).

While single monocular images provide evidence of objects in the scene -- in the form of a likelihood score for their presence, shape and pose -- they should not be used to make a decision. Instead, evidence should be accumulated over time, and the likelihood at each instant combined into a posterior estimate of object pose and identity. This is often referred to as ``semantic mapping,'' an early instance of which using depth sensors (RGB-D images) was given in~\cite{slampp}. Our method aims at the same goal, but using a monocular camera and inertial sensors, rather than a range sensor.

Inertial sensors are increasingly often present in sensor suites with monocular cameras, from cars to phones, tablets, and drones. They complement vision naturally, in an information-rich yet cheap sensor package. Unlike RGB-D, they can operate outdoor; unlike stereo, they are effective at far range; unlike lidar, they are cheap, light, and provide richer photometric signatures. Inertial sensors provide a globally consistent orientation reference (gravity) and scale up to some drift. This allows reducing pose space to four dimensions instead of six. We leverage recent developments in visual-inertial sensor fusion, and its use for semantic mapping, an early instance of which was given in~\cite{dong2017visual}, where objects were represented by bounding boxes in 3D. Our method extends that work to richer object models, that allow computing fine-grained visibility and estimating accurate pose.\\

\noindent{\bf Contributions}
We focus on applications to (indoor and outdoor) navigation, where many objects of interest are rigid and static: parked cars, buildings, furniture. 
Our contribution is a method and system that produces camera poses and a point-cloud map of the environment, populated with 3D shape and appearance models of objects recognized. It is semantic in the sense that we have identities for each object instance recognized. Also, all geometric and topological relations (proximity, visibility) are captured by this map. 

We achieve this by employing some tools from the literature, namely visual-inertial fusion, and crafting a novel likelihood model for objects and their pose, leveraging recent developments in deep learning-based object detection. The system updates its state (memory) causally and incrementally, processing only the current image rather than storing batches. 

Another contribution is the introduction of a dataset for testing visual-inertial based semantic mapping and 3D object detection. Using inertials is delicate as accurate time-stamp, calibration and bias estimates are needed. To this date, we are not aware of any dataset for object detection that comes with inertials.

We do not address intra-class variability. Having said that, the method is somewhat robust to modest changes in the model. For instance, if we have a model Aeron chair (Fig.~\ref{fig-database-groundtruth}) with arm rests, we can still detect and localize an Aeron chair without them, or with them raised or lowered.\\

\noindent{\bf Organization}
In Sect.~\ref{sect-methodology}, we describe our method, which includes top-down (filter) and bottom-up (likelihood/proposals) components. In particular, Sect.~\ref{sect-measurement} describes the novel likelihood model we introduce, using a detection and edge scoring network. Sect.~\ref{sect-implementation} describes our implementation, which is tested in Sect.~\ref{sect-experiments}, where the VISMA dataset is described. We discuss features and limitations of our method in Sect.~\ref{sect-discussion}, in relation to prior related work.

\section{Methodology}
\label{sect-methodology}
To facilitate semantic analysis in 3D, we seek to reconstruct a model of the scene sufficient to provide a Euclidean reference where to place object models. This cannot be done with a single monocular camera. Rather than using lidar (expensive, bulky), structured light (fails outdoors), or stereo (ineffective at large distances), we exploit inertial sensors frequently co-located with cameras in many modern sensor platforms, including phones and tablets, but also cars and drones. Inertial sensors provide a global and persistent orientation reference from gravity, and an estimate of scale, sufficient for us to reduce Euclidean motion to a four-dimensional group. In the next section we describe our visual-inertial simultaneous localization and mapping (SLAM) system.
\subsection{Gravity-referenced and scaled mapping}
We wish to estimate $p(Z_t, X_t|y^t)$ the joint posterior of the state of the sensor platform $X_t$ and objects in the scene $Z_t \doteq \{z\}_t^N$ given data $y^t=\{y_0, y_1, \cdots, y_t\}$ that consists of visual (image $I_t$) and inertial (linear acceleration $\alpha_t$ and rotational velocity $\omega_t$) measurements, i.e., $y_t \doteq \{I_t, \alpha_t, \omega_t\}$.  
The posterior can be factorized as 
\begin{equation}
p(Z_t, X_t|y^t) \propto p(Z_t|X_t,y^t)p(X_t|y^t)
\end{equation}
where $p(X_t|y^t)$ is typically approximated as a Gaussian distribution whose density is estimated recursively with an EKF~\cite{jazwinski70} in the visual-inertial sensor fusion literature \cite{mourikisr07,tsotsos2015robust}.
Upon convergence where the density $p(X_t|y^t)$ concentrates at the mode $\hat X_t$, the joint posterior can be further approximated using a point estimate of $\hat X_t$. 

Visual-inertial SLAM has been used for object detection by \cite{dong2017visual}, whose notation we follow here. The state of a visual-inertial sensor platform is represented as
$$X_t \doteq [\Omega_{ib}^\top, T_{ib}^\top, \Omega_{bc}^\top, T_{bc}^\top, v^\top, \alpha_\mathrm{bias}^\top, \omega_\mathrm{bias}^\top, \gamma^\top, \tau]^\top$$
where $g_{ib}(t)\doteq (\Omega_{ib}, T_{ib})\in\se$
is the transformation of the body frame to the inertial frame, $g_{bc}(t)\doteq (\Omega_{bc}, T_{bc})\in\se$ is the camera-to-body alignment, $v\in \real^3$ is linear velocity, $\alpha_\mathrm{bias}, \omega_\mathrm{bias} \in\real^3$ are accelerometer and gyroscope biases respectively, $\gamma \in\real^3$ is the direction of gravity and $\tau \in \real$ is the temporal offset between visual and inertial measurements. The transformation from camera frame to inertial frame is denoted by $g_{ic} \doteq g_{ib} g_{bc}$. The implementation details of the visual-inertial SLAM system adopted are in Sect.~\ref{sect-implementation}. Next, we focus on objects.
\subsection{Semantic Mapping}
For each object $z_t \in Z_t$ in the scene, we simultaneously estimate its pose $g \in \mathrm{SE}(3)$ and identify shape $\shape \subset \real^3$ over time.
We construct beforehand a database of 3D models, which covers objects of interest in the scene. Thus the task of estimating shape of objects is converted to the task of determining shape label $k \in \{1, 2, \cdots, K\}$ of objects, which is a discrete random variable. Once the shape label $k$ is estimated, its shape $\shape(k)$ can be simply read off from the database. Furthermore, given an accurate estimate of gravity direction $\gamma$ from visual-inertial SLAM, the 6DoF (degrees of freedom) object pose can be reduced to a four-dimensional group element $g \doteq (t, \theta)$: Translation $t \in \real^3$ and rotation around gravity (azimuth) $\theta \in [0, 2\pi)$.

We formulate the semantic mapping problem as estimating the posterior $p(z_t=\{k, g\}_t | \hat X_t, I^t)$ conditioned on mode $\hat X_t$, which can be computed in a hypothesis testing framework, of which the hypothesis space is the Cartesian product of shape label and pose $\{k\} \times \{g\}$. To facilitate computation and avoid cluttered notations, we drop $\hat X_t$ behind the condition bar and introduce an auxiliary discrete random variable: Category $c \in \{1, 2, \cdots, C\}$.
\begin{align}
		& p(\{k, g\}_{t} | I^{t}) = \sum_{c_{t}} p(\{k, g, c\}_{t} | I^{t}) \\
		& \propto \sum_{c_{t}}                                              
	p(I_{t}| \{k, g, c\}_{t})
	\int p(\{k, g, c\}_{t} | \{k, g, c\}_{t-1}) dP(\{k, g, c\}_{t-1}|I^{t-1})
\end{align}
where marginalization is performed over all possible categories. By noticing that category $c_{t}$ is a deterministic function of shape label $k_{t}$, i.e. $p(c_t|k_t)=\delta(c_t-c(k_t))$, the posterior $p(\{k, g\}_t|I^t)$ can be further simplified as follows:
\begin{equation}
	\sum_{c_{t}} \delta(c_t-c(k_t))
	{p(I_{t}| \{k, g, c\}_{t})}{\int p(\{k, g\}_{t} | \{k, g\}_{t-1}) dP(\{k, g\}_{t-1}|I^{t-1})}
	\label{eq-posterior}
\end{equation}
where the first term in the summation is the likelihood (Sect.~\ref{sect-measurement}) and the second term can be approximated by numerical integration of weighted particles (Sect.~\ref{sect-bootstrap}).
\subsection{Parameterization and Dynamics}
\label{sect-pose-parametrization}
Each object is parametrized locally and attached to a reference camera frame at time $t_r$ with pose $g_{ic}(t_r)$ and the translational part of object pose is parameterized by a bearing vector $[x_c, y_c]^\top\in\real^2$ in camera coordinates and a log depth $\rho_c \in\real$ where $z_c=\exp(\rho_c)\in\real_+$. Log depth is adopted because of the positivity and cheirality it guarantees. Inverse depth~\cite{civera2008inverse}, though often used by the SLAM community, has singularities and is not used in our system.
The object centroid is then $T_{co}=\exp(\rho_c)\cdot[x_c, y_c, 1]^\top$ in the reference camera frame and $T_{io}=g_{ic}(t_r)T_{co}$ in the inertial frame.
For azimuth $\theta$, we parameterize it in the inertial frame and obtain the rotation matrix via Rodrigues' formula:
$$
\mathbf{R}_{io}(\theta)=\mathbf{I}+\sin\theta\widehat\gamma+(1-\cos\theta)\widehat\gamma^2
$$
where $\gamma$ is the direction of gravity and the hat operator $\widehat{\cdot}$ constructs a skew-symmetric matrix from a vector.
Therefore the object pose in the inertial frame is $g_{io}=[\mathbf{R}_{io} |T_{io}] \in\se$.
Although the pose parameters are unknown constants instead of time varying quantities, we treat them as stochastic processes with trivial dynamics as a common practice: $[\dot x_c, \dot y_c, \dot\rho_c, \dot \theta]^\top=[n_x, n_y, n_\rho, n_\theta]^\top$
where $n_x, n_y, n_\rho$ and $n_\theta$ are zero-mean Gaussian noises with small variance.
\subsection{Measurement Process}
\label{sect-measurement}
In this section, we present our approximation to the log-likelihood $L(\{k, g, c\}_t| I_t) \doteq \log p(I_t|\{k, g, c\}_t)$ of the posterior~\eqref{eq-posterior}. Given the prior distribution $p(\{k, g\}_{t-1}|I^{t-1})$, a hypothesis set $\{k, g\}_t$ can be constructed by a diffusion process around the prior $\{k,g\}_{t-1}$.
To validate the hypothesis set, we use a log-likelihood function which consists of two terms:
\begin{equation}
	L(\{k, g, c\}_t|I_t)=\alpha\cdot\Phi_\text{CNN}(\{k, g, c\}_t|I_t) + \beta\cdot\Phi_\text{edge}(\{k, g\}_t|I_t)
\end{equation}
where $\alpha$ and $\beta$ are tuning parameters.
The first term in the log-likelihood is a convolutional neural network which measures the likelihood of an image region  is to contain a certain object. The second term scores the likelihood of an edge in the image. We describe them in order.
\subsubsection{CNN as Likelihood Mechanism}
Given a hypothesis $\{k, g\}_t$ in the reference frame, we first bring it to the current camera frame by applying a relative transformation and then project it to the current image plane via a rendering process. A minimal enclosing bounding box of the projection is found and then fed into an object detection network. The score of the hypothesis is simply read off from the network output (Fig.~\ref{fig-flowchart}).
\begin{equation}
	\Phi_\text{CNN}(k, g, c; I)=\mathrm{Score}\Big(I_{|b=\pi\big( g_{ic}^{-1}(t)g_{io}(t_r)\shape(k) \big)}, c\Big)
\end{equation}
where $\pi(\cdot)$ denotes the process to render the contour map of the object of which the minimal enclosing bounding box $b$ is found; $g_{io}(t_r)$ is the transformation to bring the object from local reference frame at time $t_r$ to the inertial frame and $g_{ic}^{-1}(t)$ is the transformation to bring the object from the inertial frame to current camera frame.

Either a classification network or a detection network can be used as our scoring mechanism. However, due to the size of the hypothesis set at each time instant, which is then mapped to bounding boxes sitting on the same support, it is more efficient to use a detection network where the convolutional features are shared by object proposals via ROI pooling: Once predicted, all the box coordinates are fed to the second stage of Faster R-CNN as object proposals in a single shot, where only one forward pass is carried out.
\subsubsection{Edge likelihood}
\label{sect-edge-likelihood}
An object detection network is trained to be invariant to viewpoint change and intra-class variabilities, which makes it ill-suited for pose estimation and shape identification. To that end,  we train a network to measure the likelihood of edge correspondence:
\begin{equation}
	\Phi_\text{edge}(k, g; I)=h\Big(\pi\big(g_{ic}^{-1}(t)g_{io}(t_r)\shape(k)\big), \mathrm{EdgeNet}(I)\Big)
\end{equation}
where $h(\cdot, \cdot)$ is some proximity function which measures the proximity of edge map constructed from pose and shape hypothesis via rendering (first argument of $h$) and edge map extracted from the image (second argument of $h$). 
	
A popular choice for proximity function $h$ is one-dimensional search~\cite{blake1997condensation,drummond2002real,klein2006full}, which we adopt (see Sup.~Mat. for details). Such a method is geometric and more robust than appearance based methods which are photometric and subject to illumination change. However, due to its nature of locality, this method is also sensitive to background clutter and can be distracted by texture-rich image regions. Fortunately, these weaknesses are easily compensated by $\Phi_\text{CNN}$ which has a large receptive field and is trained on semantics. Also, instead of using Canny~\cite{canny1987computational} or other non-learning-based edge features, we design an edge detection network~(Sect.~\ref{sect-modules}) on semantically relevant training sets. Fig.~\ref{fig-scenenn-visual} shows examples illustrating background distraction.
	
\section{Implementation Details}
\label{sect-implementation}
\begin{figure}
	\vspace{-10pt}
	\begin{center}
			\subfloat[\scriptsize System flowchart]{\includegraphics[width=0.4\linewidth]{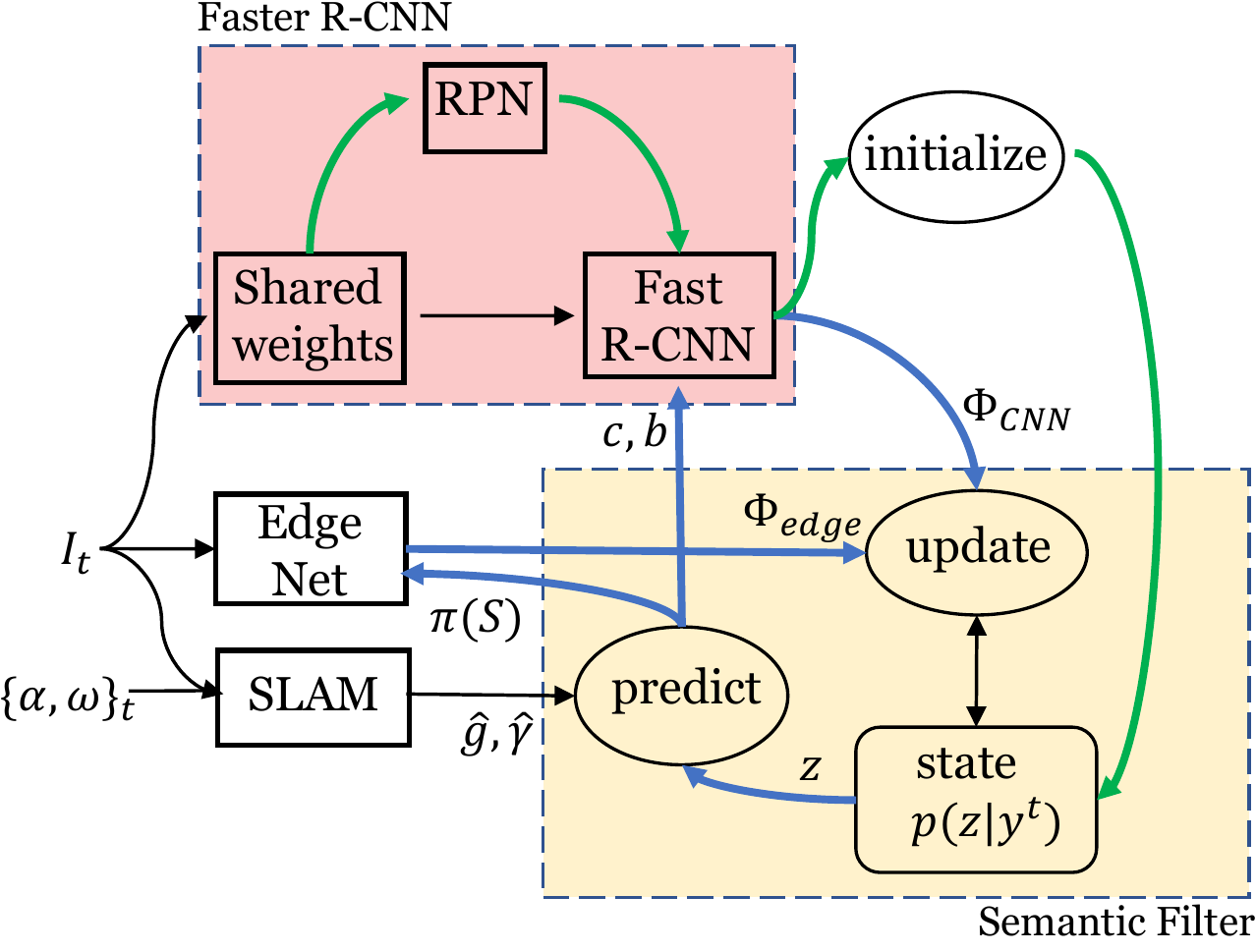}
			\label{fig-flowchart}}
			\hspace{+20pt}
			\subfloat[\scriptsize Scoring process]{\includegraphics[width=0.4\linewidth]{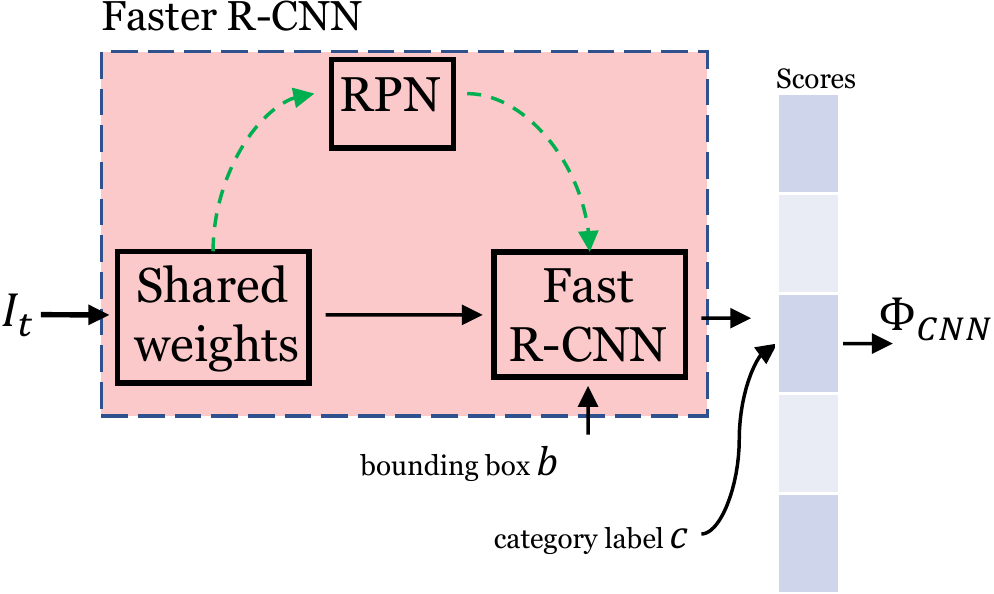}
			\label{fig-scoring}}
	\end{center}
	\vspace{-15pt}
	\caption{\captionfont \textbf{Left} \textit{System flowchart.} Green pathway: Faster R-CNN as a bottom-up proposal generation mechanism. Blue pathway: Top-down hypothesis validation process. Pink box: Faster R-CNN. Yellow box: Semantic filter.
	\textbf{Right} \textit{CNN as scoring mechanism.} 
	Dashed pathway (proposal generation) is inactive during hypothesis testing. See system overview of Sect.~\ref{sect-overview} for details.}
	\label{fig-flowchart}
	\vspace{-15pt}
\end{figure}

\noindent{\bf System Overview}
\label{sect-overview}
An overview of the system is illustrated in the system flowchart (Fig.~\ref{fig-flowchart}). We perform Bayesian inference by interleaving bottom-up (the green pathway) and top-down (the blue pathway) processing over time, which both rely on CNNs. Faster R-CNN as a bottom-up proposal generation mechanism takes input image $I_t$ and generates proposals for initialization of new objects. In the top-down hypothesis validation process, both geometric (edge net, takes object contour $\pi(S)$ and outputs likelihood $\Phi_\text{edge}$) and semantic (Fast R-CNN, takes predicted bounding box $b$ and class label $c$ and outputs likelihood $\Phi_\text{CNN}$) cues are used. 
Faster R-CNN consists of a region proposal network (RPN) and a Fast R-CNN, which share weights at early convolutional layers. RPN is only activated in the bottom-up phase to feed Fast R-CNN object proposals of which bounding box coordinates are regressed and class label is predicted. During top-down phase, proposals needed by Fast R-CNN are generated by first sampling from the prior distribution $p(z|y^{t-1})$ followed by a diffusion and then mapping each sample to a bounding box $b$ and a class label $c$. Fig.~\ref{fig-scoring} illustrates the scoring process. 
The semantic filter (yellow box) is a variant of bootstrap algorithm~\cite{gordon1993novel} and recursively estimates the posterior $p(z|y^t)$ as a set of weighted particles. Point estimates of gravity $\gamma$ and camera pose $g$ are from the SLAM module.

\noindent{\bf SLAM and Network Modules}
\label{sect-modules}
We implement the system in C++ and OpenGL Shading Language (GLSL, for rendering) and follow a modular design principle: Each major module runs in its own process and communicates via a publish/subscribe message transport system, which enables expandability and possible parallelism in the future. The visual-inertial SLAM is based on~\cite{tsotsos2015robust} which produces gravity-referenced and scaled camera pose estimates needed by the semantic mapping module. An off-the-shelf Faster R-CNN implementation~\cite{detectron2018} with weights pre-trained on Microsoft COCO is turned into a service running constantly in the background. Note we take the most generic object detector as it is {\it without fine-tuning on specific object instances}, which differs from other object instance detection systems. The benefit is scalability: No extra training is required when novel object instances are spotted. For the weakly semantic-aware edge detection network, we adapt SegNet~\cite{segnet} to the task of edge detection: The last layer of SegNet is modified to predict the probability of each pixel being an edge pixel. Weights pre-trained on ImageNet are fine-tuned on BSDS~\cite{BSDS}. Fig.~\ref{fig-visual} shows sample results of our edge detection network.\\
\noindent{\bf Occlusion and Multiple Objects}
\label{sect-occlusion}
We turn to some heuristics to handle occlusion due to its combinatorial nature. Fortunately, this is not a problem because we explicitly model the shape of objects, of which a Z-Buffer of the scene can be constructed with each object represented as its most likely shape at expected pose (Fig.~\ref{fig-visual} and~\ref{fig-scenenn-visual}). Only the visible portion of the edge map is used to measure the edge likelihood while Faster R-CNN still runs on the whole image, because object detectors should have seen enough samples with occlusion during the training phase and thus robust to occlusion.\\
\noindent{\bf Initialization} 
\label{sect-initialization}
An object proposal from Faster R-CNN is marked as ``explained'' if it overlaps with the predicted projection mask by a large margin. For those ``unexplained'' proposals, we initialize an object attached to the current camera frame by spawning a new set of particles. For each particle: 
The bearing vector $[x_c, y_c]^\top$ is initialized as the direction from the optical center to the bounding box center with a Gaussian perturbation. 
The log depth is initialized at a nominal depth value with added Gaussian noise. 
Both the azimuth and the shape label are sampled from uniform priors. More informative priors enabled by data-driven approaches are left for future investigation.\\
\noindent{\bf The Semantic Filter}
\label{sect-bootstrap}
We summarize our joint pose estimation and shape identification algorithm in Alg.~\ref{alg-bootstrap}, which is a hybrid bootstrap filter~\cite{gordon1993novel} with Gaussian kernel for dynamics and a discrete proposal distribution for shape identification: The shape label stays the same with high probability and jumps to other labels equally likely to avoid particle impoverishment. A breakdown of the computational cost of each component can be found in the Sup.~Mat.
\vspace{-15pt}
\begin{algorithm}
	\caption{\small Semantic Filter}
	\begin{algorithmic}
		\begin{small}
			\STATE \textbf{1. Initialization}\\
			When an unexplained bottom-up proposal is found at time $t=t_r$, sample $\{k, g\}_{t_r}^{(i)} \sim p(\{k, g\}_{t_r})$ and attach object to camera frame $t_r$. (Sect.~\ref{sect-initialization}, Initialization)
			\STATE \textbf{2. Importance Sampling}\\
			At time $t \ge t_r$, sample $\{k, g\}_t^{(i)} \sim q(k_t^{(i)}|k_{t-1}^{(i)})\mathcal{N}(g_t^{(i)};g_{t-1}^{(i)}, \Sigma_{t-1})$
			and compute weights $w_t^{(i)}=\exp\big( \alpha \cdot \Phi_\text{CNN} + \beta \cdot \Phi_\text{edge}\big)$. (Sect.~\ref{sect-measurement})
			\STATE \textbf{3. Resampling}\\
			Resample particles $\{k, g\}_t^{(i)}$ with respect to the normalized importance weights $w_t^{(i)}$ to obtain equally weighted particles $\{k, g\}_t^{(i)}$.
			\STATE \textbf{4. Occlusion handling}\\
			Construct Z-Buffer at mean state to explain away bottom-up object proposals. (Sect.~\ref{sect-occlusion}, Occlusion)\\
			Set $t \leftarrow t+1$ and go to step 1.
		\end{small}
	\end{algorithmic}
	\label{alg-bootstrap}
\end{algorithm}

\section{Experiments}
\label{sect-experiments}
We evaluate our system thoroughly in terms of mapping and object detection. While there are several benchmarks for each domain, very few allow measuring simultaneously localization and reconstruction accuracy, as well as 3D object detection. 

In particular, \cite{sturm2012benchmark,handa2014benchmark} are popular for benchmarking RGB-D SLAM: one is real, the other synthetic. KITTI~\cite{geiger2013vision} enables benchmarking SLAM as well as object detection and optical flow. Two recent visual-inertial SLAM benchmarks are~\cite{burri2016euroc} and~\cite{pfrommerSDC17}. Unfortunately, we find these datasets unsuitable to evaluate the performance of our system: Either there are very few objects in the dataset~\cite{sturm2012benchmark,handa2014benchmark,burri2016euroc,pfrommerSDC17}, or there are many, but no ground truth shape annotations are available~\cite{geiger2013vision}.

On the other hand, object detection datasets~\cite{pascalvoc,imagenet,microsoftcoco} focus on objects as regions of the image plane, rather than on the 3D scene. \cite{xiang_wacv14,objectnet3d} are among the few exploring object attributes in 3D, but are single-image based. Not only does our method leverage video imagery, but it requires a Euclidean reference, in our case provided by inertial sensors, making single-image benchmarks unsuitable.

Therefore, to measure the performance of our method, we had to construct a novel dataset, aimed at measuring performance in visual-inertial semantic mapping. We call this the VISMA set, which will be made publicly available upon completion of the anonymous review process, together with the implementation of our method.

\begin{figure}
 \centering
 \begin{tabular}{ccc}
    \includegraphics[width=0.31\textwidth]{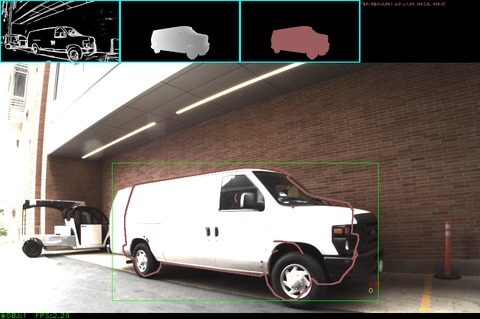} &
    \includegraphics[width=0.31\textwidth]{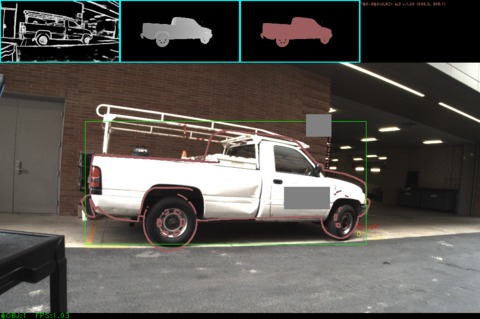} &
    \includegraphics[width=0.3\textwidth]{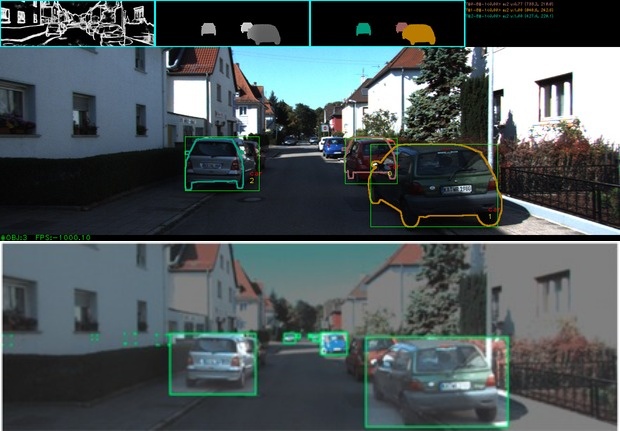} \\
 \end{tabular}
 \vspace{-5pt}
 \caption{\captionfont \textit{Exemplary outdoor results.} (best in color at $5 \times$) In each panel, top inset shows (left to right): edge map, Z-buffer, projection masks; bottom shows input RGB with predicted mean object boundary and CNN detection. Rightmost panel shows a visual comparison of ours (top) against Fig. 1 of ~\cite{dong2017visual} (bottom), where we capture the boundaries of the cars better. Though only generic models from ShapeNet are used in these examples, pose estimates are fairly robust to shape variations. }
 \label{fig-outdoor}
 \vspace{-15pt}
\end{figure}

VISMA contains 8 richly annotated videos of several office scenes with multiple objects, together with time-stamped inertial measurements. We also provide ground truth annotation of several objects (mostly furniture, such as chairs, couches and tables) (Sect.~\ref{sect-groundtruth}). Over time we will augment the dataset with additional scanned objects, including moving ones, and outdoor urban scenes. The reason for selecting indoors at first is because we could use RGB-D sensors for cross-modality validation, to provide us with pseudo-ground truth. Nevertheless, to demonstrate the outdoor-applicability of our system, we provide illustrative results on outdoor scenes in Fig.~\ref{fig-outdoor}.

We also looked for RGB-D benchmarks and datasets, where we could compare our performance with independently quantified ground truth. SceneNN~\cite{hua2016scenenn} is a recently released RGB-D dataset, suitable for testing at least the semantic mapping module of our system, even though originally designed for deep learning. Sect.~\ref{sect-scenenn} describes the experiments conducted on SceneNN.

\subsection{VISMA Dataset}
\label{sect-database}
A customized sensor platform is used for data acquisition: An inertial measurement unit (IMU) is mounted atop camera equipped with a wide angle lens. The IMU produces time-stamped linear acceleration and rotational velocity readings at $100 \mathrm{Hz}$. The camera captures $500\times 960$ color images at $30 \mathrm{Hz}$. We have collected $8$ sequences in different office settings, which cover $\sim 200 \mathrm{m}$ in trajectory length and consist of $\sim 10\mathrm{K}$ frames in total. 

\begin{figure}
\vspace{-20pt}
 \centering
 \subfloat{
    \includegraphics[width=0.18\linewidth]{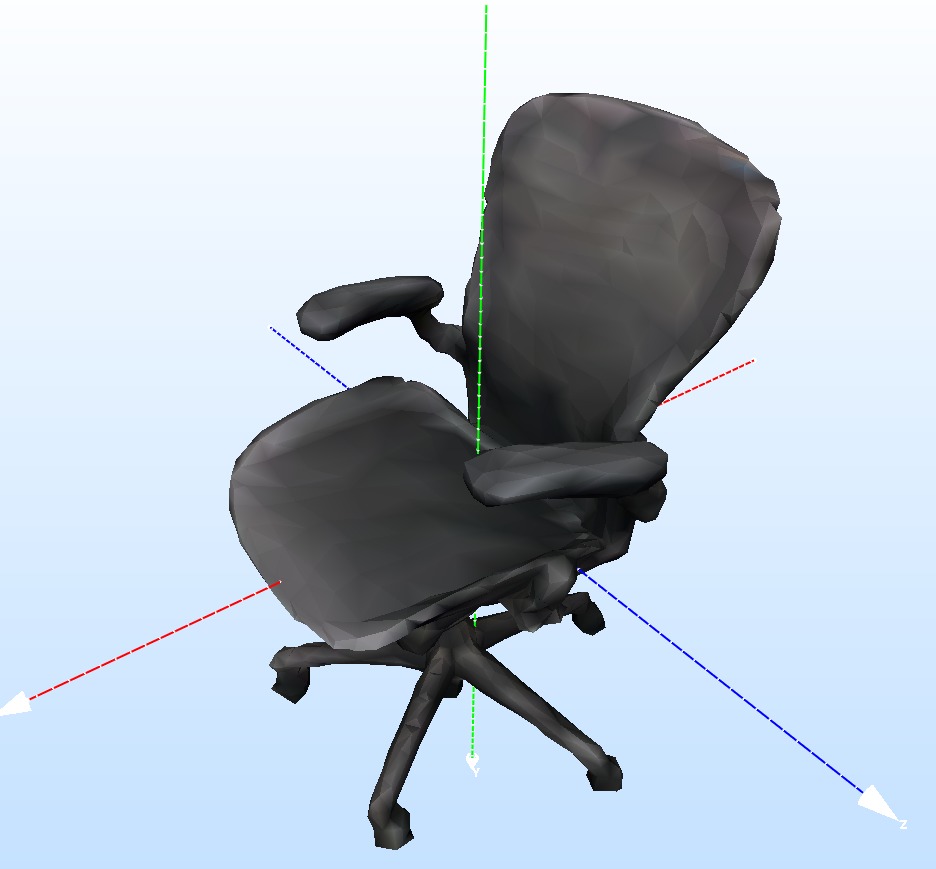}
 }
 \subfloat{
    \includegraphics[width=0.18\linewidth]{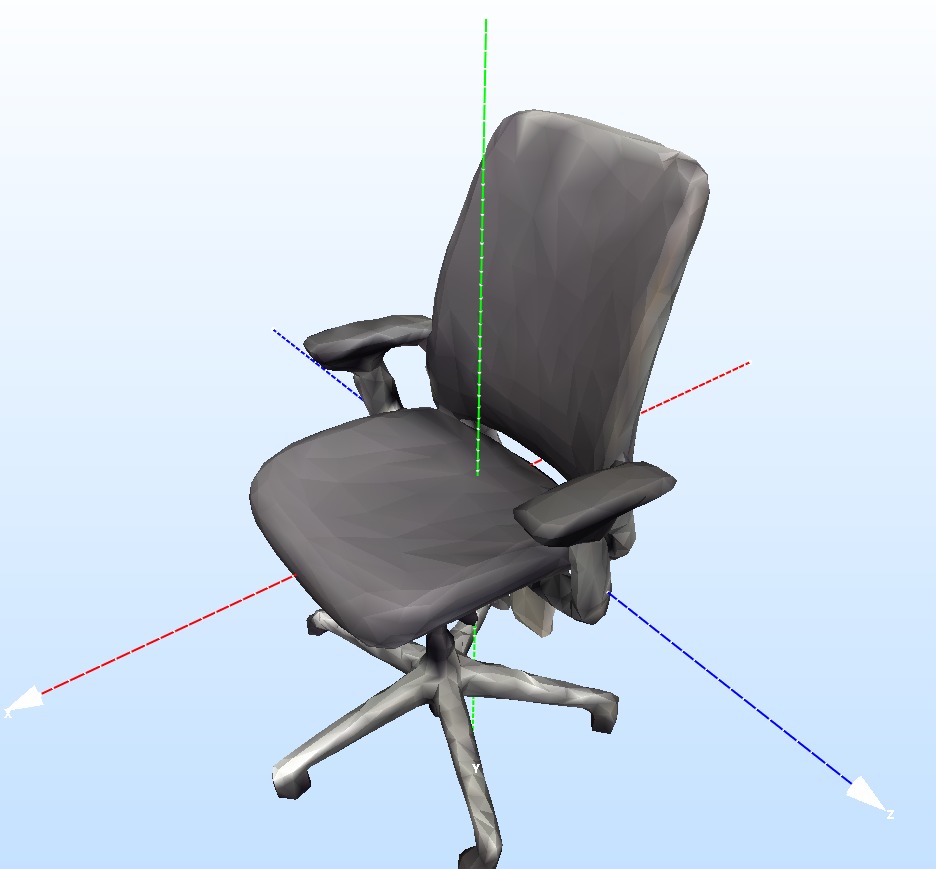}
 }
 \subfloat{
    \includegraphics[width=0.18\linewidth]{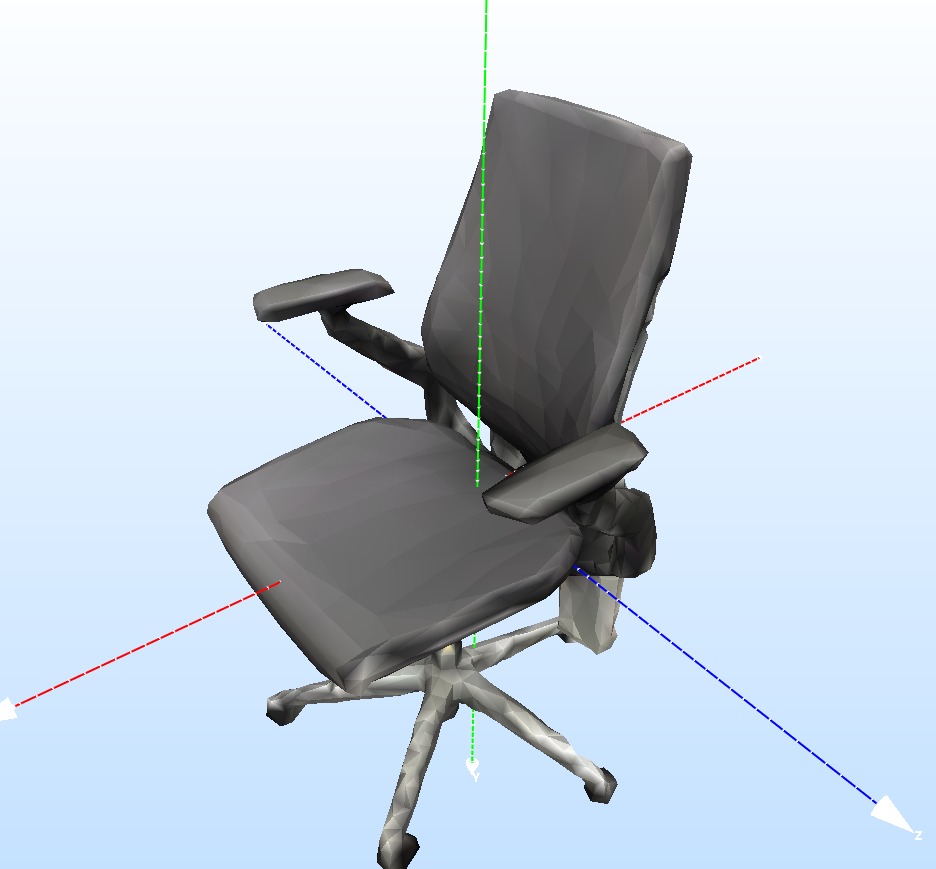}
 }
 \subfloat{
    \includegraphics[width=0.18\linewidth]{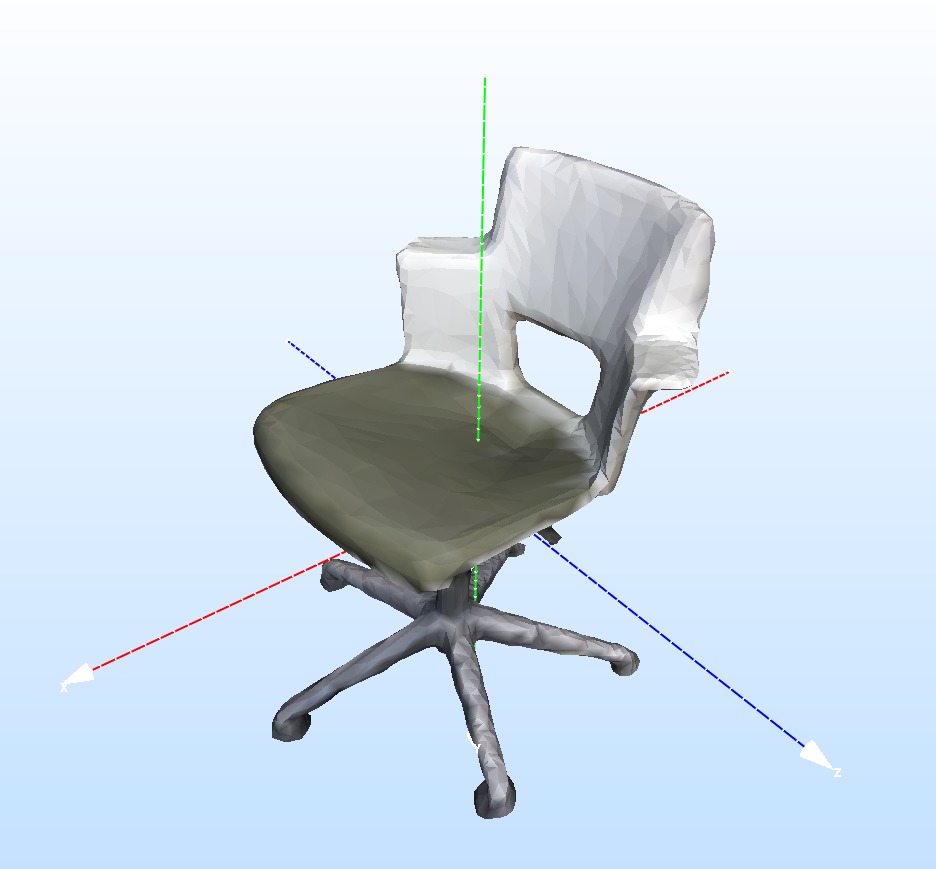}
 }
 \subfloat{
    \includegraphics[width=0.18\linewidth]{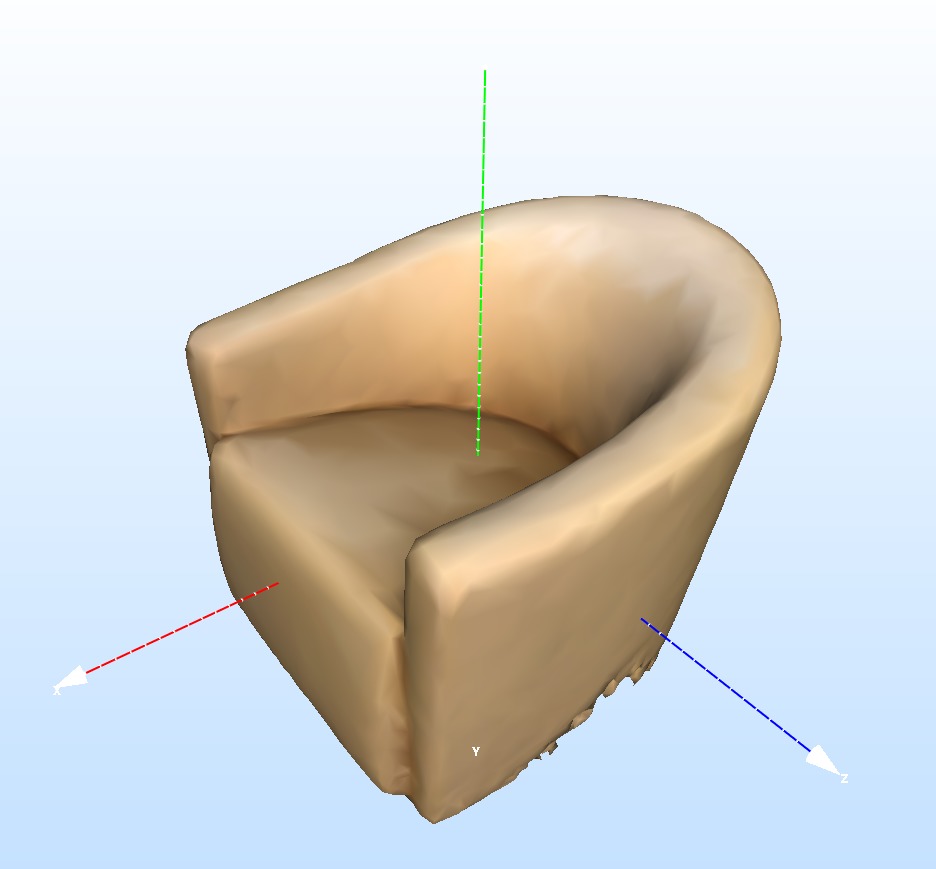}
 }\\
 \vspace{-5pt}
 \subfloat{
    \includegraphics[width=0.3\linewidth]{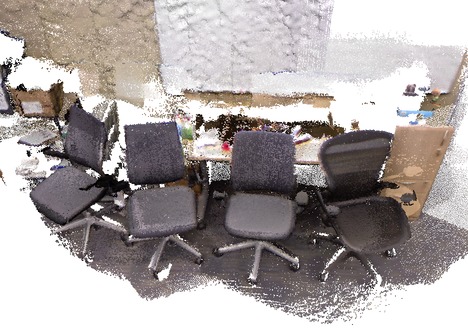}
 }
 \subfloat{
    \includegraphics[width=0.33\linewidth]{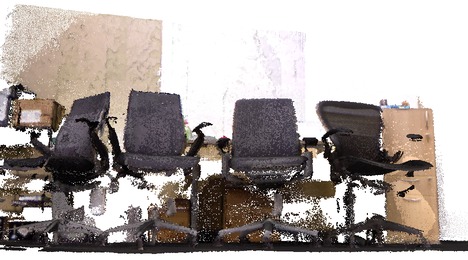}
 }
 \subfloat{
    \includegraphics[width=0.3\linewidth]{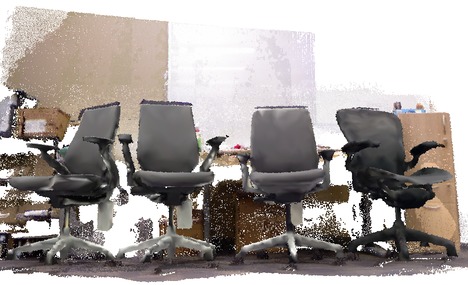} 
 }
 \caption{\captionfont 
 \textbf{Top} \textit{Sample objects in the VISMA dataset.} Each mesh has $\sim $5000 faces and is placed in an object-centric canonical frame, simplified, and texture-mapped. 
 \textbf{Bottom} \textit{(Pseudo) ground truth} from different viewpoints with the last panel showing an augmented view with models aligned to the original scene.}
 \label{fig-database-groundtruth}
 \vspace{-20pt}
\end{figure}

To construct the database of 3D models, we rely on off-the-shelf hardware and software, specifically an Occipital Structure Sensor~\footnote{\url{http://www.structure.io}} on an iPad, to reconstruct furniture objects in office scenes with the built-in 3D scanner application. This is a structured light sensor that acts as an RGB-D camera to yield water-tight surfaces and texture maps. We place the 3D meshes in an object-centric canonical frame and simplify the meshes via quadratic edge collapse decimation using MeshLab~\footnote{\url{http://www.meshlab.net}}. Top row of Fig.~\ref{fig-database-groundtruth} shows samples from our database. While the database will eventually be populated by numerous shapes, we use a small dictionary of objects in our experiments, following the setup of~\cite{slampp}. An optional shape retrieval~\cite{shrec} process can be adopted for larger dictionaries, but this is beyond the scope of this paper and not necessary given the current model library.

\subsection{Evaluation}
\label{sect-quantitative}
Comparing dense surface reconstruction is non-trivial, and several approaches have been proposed for RGB-D SLAM: Sturm et al.~\cite{sturm2012benchmark} use pose error (RPE) and absolute trajectory error (ATE) to evaluate RGB-D odometry. To ease the difficulty of ground truth acquisition, Handa et al.~\cite{handa2014benchmark} synthesized a realistic RGB-D dataset for benchmarking both pose estimation and surface reconstruction, according to which, the state of the art RGB-D SLAM systems have typical ATE of $1.1 \sim 2.0 \text{cm}$ and average surface error of $0.7\sim 2.8 \text{cm}$~\cite{whelan2015elasticfusion}, which renders RGB-D SLAM a strong candidate as our (pseudo) ground truth for the purpose of evaluating visual-inertial-semantic SLAM system.\\

\noindent{\bf Ground Truth}
\label{sect-groundtruth}
To obtain (pseudo) ground truth reconstruction of experimental scenes, we run ElasticFusion~\cite{whelan2015elasticfusion}, which is at state-of-the-art in RGB-D SLAM, on data collected using a Kinect sensor. In cases where only partial reconstruction of objects-of-interest was available due to failures of ElasticFusion, we align meshes from our  database to the underlying scene via the following procedure: Direction of gravity is first found by computing the normal to the ground plane which is manually selected from the reconstruction. Ground truth alignment of objects is then found by rough manual initialization followed by orientation-constrained ICP~\cite{Zhou2018} where only rotation around gravity is allowed. 
Bottom row of Fig.~\ref{fig-database-groundtruth} shows a reconstructed scene from different viewpoints where the last panel shows an augmented view.\\

\begin{figure}
	\centering
	\begin{tabular}{ccc}
		\includegraphics[width=0.3\linewidth]{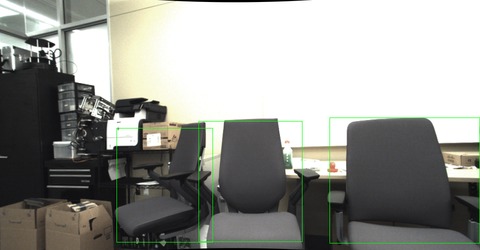}&   
		\includegraphics[width=0.3\linewidth]{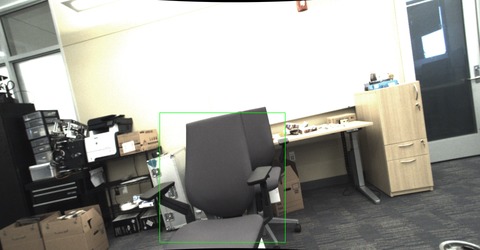}&   
		\includegraphics[width=0.3\linewidth]{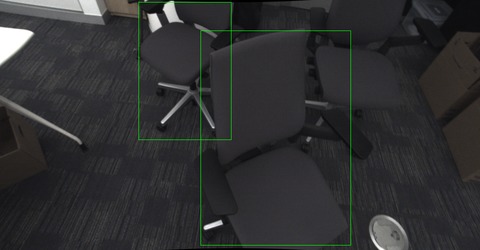}\\
		\includegraphics[width=0.3\linewidth]{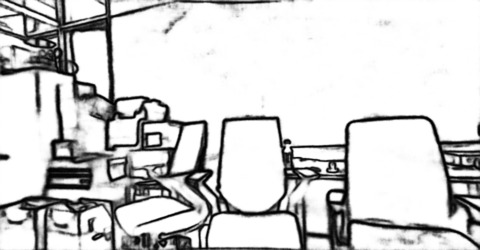}&   
		\includegraphics[width=0.3\linewidth]{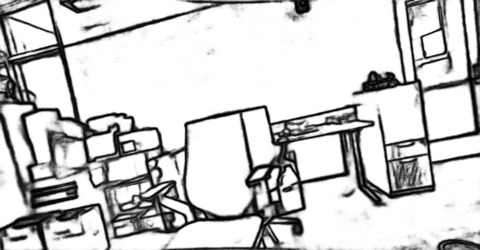}&   
		\includegraphics[width=0.3\linewidth]{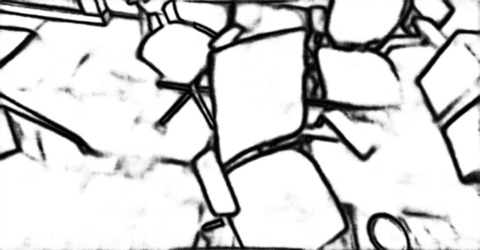}\\
		\includegraphics[width=0.3\linewidth]{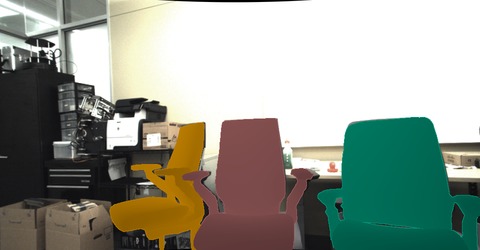}&   
		\includegraphics[width=0.3\linewidth]{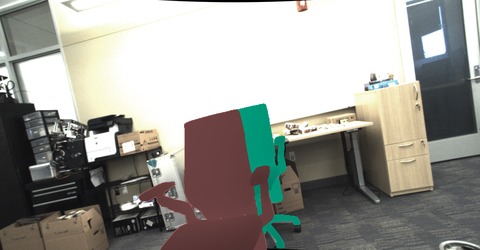}&   
		\includegraphics[width=0.3\linewidth]{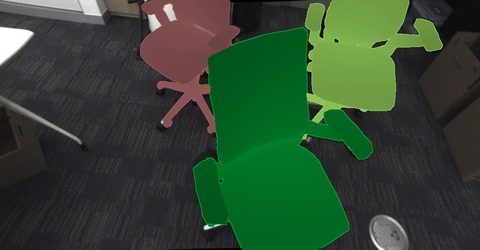}\\
		\includegraphics[width=0.3\linewidth]{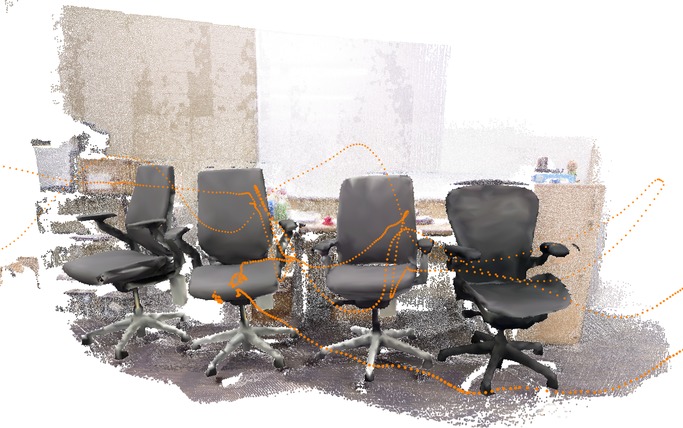}&   
		\includegraphics[width=0.3\linewidth]{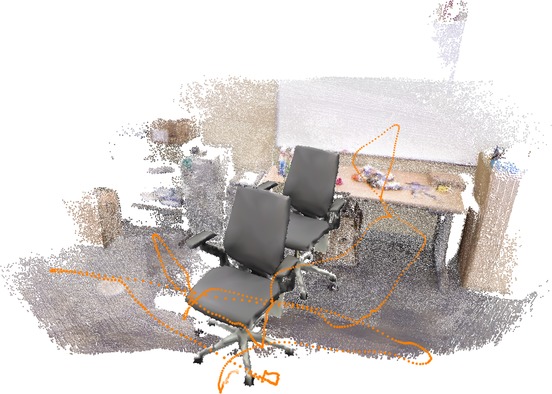}&   
		\includegraphics[width=0.3\linewidth]{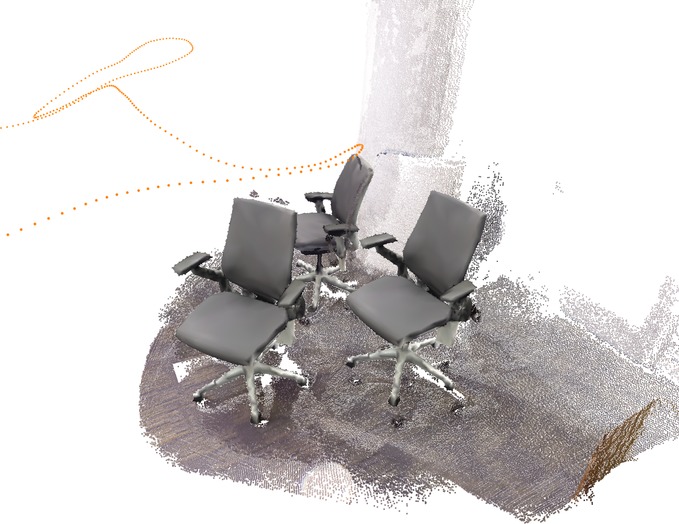}\\
		\includegraphics[width=0.3\linewidth]{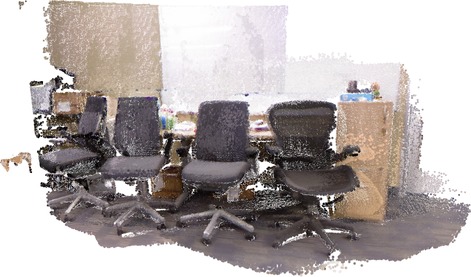}&   
		\includegraphics[width=0.3\linewidth]{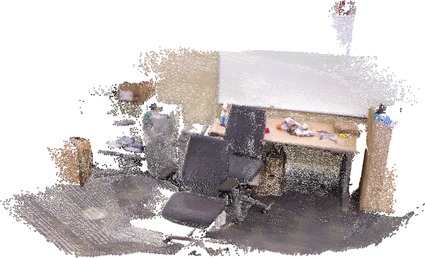}&   
		\includegraphics[width=0.3\linewidth]{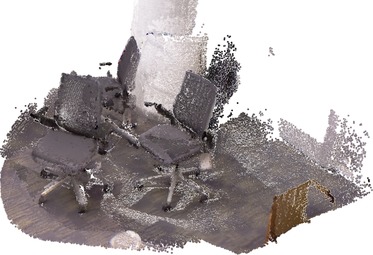}\\
		\scriptsize{Clutter2} & \scriptsize{Occlusion1} & \scriptsize{Occlusion2} 
	\end{tabular}
	\caption{\captionfont {\it Qualitative results.} (best in color at $5 \times$) Each column shows (top to bottom): One frame of the input video with CNN bounding box proposals with confidence $>0.8$; Extracted edge map; Frame overlaid with predicted instance masks shaded according to Z-Buffer -- darker indicates closer; Background reconstruction augmented with camera trajectory (orange dots) and semantic reconstruction from our visual-inertial-semantic SLAM; Ground truth dense reconstruction. Missed detections due to heavy occlusion (middle column) and indistinguishable background (right column) are resolved by memory and inference in a globally consistent spatial frame.}
	\label{fig-visual}
\end{figure}

\noindent{\bf Metrics and Results}
We adopt the surface error metric proposed by~\cite{handa2014benchmark} for quantitative evaluation. First, a scene mesh is assembled by retrieving 3D models from the database according to the most likely shape label, to which the pose estimate is applied. A point cloud is then densely sampled from the scene mesh and aligned to the ground truth reconstruction from RGB-D SLAM via ICP, because both our reconstructed scene and the ground truth scene are up to an arbitrary rigid-body transformation. Finally, for each point in the aligned scene mesh, the closest triangle in the ground truth scene mesh is located and the normal distance between the point and the closest triangle is recorded.
Following~\cite{handa2014benchmark}, four standard statistics are computed over the distances for all points in the scene mesh: Mean, median, standard deviation, and max (Table~\ref{tab-surface-error}). In addition to surface error, Table~\ref{tab-surface-error} also includes pose estimation error which consists of translational and rotational part. Fig.~\ref{fig-visual} shows how common failures of an image-based object detector have been resolved by memory (state of the semantic filter) and inference in a globally consistent spatial frame.
\setlength{\tabcolsep}{4pt}
\begin{table}[]
\vspace{-20pt}
\centering
\caption{\captionfont {\it Surface error and pose error} measured over 4 sequences from the VISMA dataset. Qualitative results on the other 4 sequences with coarse annotations can be found in the Sup.~Mat.
Translational error reads $\| T_{gt}-\hat T\|_2$ and rotational error reads $\|\log^\vee(\hat R^\top R_{gt})\|_2$, where $\log : \mathrm{SO}(3)\mapsto \mathrm{so}(3)$ and $^\vee: \mathrm{so}(3) \mapsto \mathbb R^3$. $(R_{gt}, T_{gt})$ and $(\hat R, \hat T)$ are ground truth and estimated object pose respectively.}
\label{tab-surface-error}
\begin{\tablefontsize}
\begin{tabular}{ll|cccc}
\hline 
\multicolumn{2}{l|}{Error Metric}  & Clutter1 & Clutter2 & Occlusion1 & Occlusion2\\
\hline
\multirow{4}{1cm}{Surface} & Median(cm)   & $1.37$ & $1.11$  & $1.30$ & $2.01$ \\
& Mean(cm)     & $1.99$ & $1.39$  & $1.73$ & $2.79$ \\
& Std.(cm)     & $1.96$ & $1.12$  & $1.45$ & $2.54$ \\
& Max(cm)      & $17.6$ & $9.88$  & $14.3$ & $17.9$ \\
\hline
\multirow{2}{1cm}{Pose} & Mean Trans. (cm) & 4.39 & 2.42 & 3.94 & 13.64 \\
& Mean Rot. (degree) & 6.16 & 4.66 & 4.86 & 9.12 \\
\hline
\end{tabular}
\end{\tablefontsize}
\end{table}
\setlength{\tabcolsep}{1.4pt}

\subsection{Experiments on SceneNN Dataset}
\label{sect-scenenn}
For independent validation, we turn to recent RGB-D scene understanding datasets to test at least the semantic mapping part of our system.
Although co-located monocular and inertial sensors are ubiquitous, hence our choice of sensor suite, any SLAM alternative can be used in our system as the backbone localization subsystem as long as reliable metric scale and gravity estimation are provided. This makes SceneNN suitable for testing the semantic mapping part of our system, although originally designed for RGB-D scene understanding. It provides ground truth camera trajectories in a gravity-aligned reference frame. Raw RGB-D streams and ground truth meshes reconstructed from several object-rich real world scenes are provided in SceneNN. 

To test the semantic mapping module on SceneNN, we take the ground truth camera trajectory and color images as inputs. {\it Note the depth images are not used in our experiments.} The database is constructed by manually selecting and cropping object meshes from the ground truth scene mesh. A subset scenes of SceneNN with various chairs is selected for our experiments.
Except the fact that the camera trajectory and gravity are from the ground truth instead of from our visual-inertial SLAM, the rest of the experiment setup are the same as those in the experiment on our own dataset. Table~\ref{tab-scenenn-surface-error} shows statistics of surface error of our semantic mapping on SceneNN. Typical mean surface error is around $3 \mathrm{cm}$. Fig.~\ref{fig-scenenn-visual} shows some qualitative results on SceneNN.
\setlength{\tabcolsep}{2.5pt}
\begin{table}
\vspace{-20pt}
\captionsetup[subtable]{labelformat=empty}
\centering
\begin{scriptsize}
\caption{\captionfont {\it Surface error} measured on a subset of the SceneNN dataset.}
\label{tab-scenenn-surface-error}
\begin{tabular}{|l|c|c|c|c|c|c|c|c|c|c|c|c|c|c|c|}
\hline 
Sequence   & 005  & 025   & 032  & 036  & 043  & 047  & 073  & 078  & 080  & 082  & 084  & 096   & 273  & 522 & 249 \\
\hline                                                
Median(cm) & 1.84 & 0.726 & 3.08 & 2.25 & 3.66 & 3.10 & 2.59 & 3.04 & 2.82 & 2.35 & 1.29 & 0.569 & 2.06 & 1.31 & 0.240 \\
Mean(cm)   & 3.47 & 0.756 & 6.28 & 4.10 & 4.24 & 4.11 & 3.04 & 3.51 & 3.15 & 3.32 & 1.70 & 0.684 & 2.15 & 1.69 & 0.299\\
Std.(cm)   & 3.48 & 0.509 & 6.95 & 5.10 & 3.11 & 3.52 & 2.17 & 2.60 & 2.09 & 2.99 & 1.51 & 0.518 & 1.24 & 1.39 & 0.217\\
Max(cm)    & 13.7 & 3.07  & 36.3 & 34.3 & 11.9 & 18.5 & 8.72 & 17.4 & 13.9 & 22.7 & 8.33 & 4.41  & 5.75 & 5.60 & 1.27\\
\hline
\end{tabular}
\end{scriptsize}
\vspace{-20pt}
\end{table}
\setlength{\tabcolsep}{1.4pt}

\section{Discussion}
\label{sect-discussion}
Our method exploits monocular images and time-stamped inertial measurements to construct a point-cloud model of the environment, populated by object models that were recognized, along with the camera trajectory in an Euclidean frame. We target indoor and outdoor mobility scenarios, and focus on indoor for evaluation due to the availability of benchmark. Yet no benchmark has inertial and semantic ground truth, so we have introduced VISMA.

We believe most mapping and navigation methods in the near future will utilize this modality as it is ubiquitous (e.g., in every smart phone or  car, even some vacuum cleaners). Yet, at present, ours is one of few methods to exploit inertials for semantic mapping in the literature. 

Our method has several limitations: It is limited to rigid objects and static scenes; it is susceptible to failure of the low-level processing modules, such as the detection or edge networks. It works for object instances, but cannot handle intra-class variability. It is not operating in real time at present, although it has the potential to. 

Future extensions of this work include expansions of the VISMA dataset, the addition of synthetic scenes with rich ground truth. Extensions to independently moving objects, and deforming objects, is also an open area of investigation.\\

\begin{figure}[t]
	\begin{center}
	\begin{tabular}{cccc}
		\includegraphics[width=0.22\linewidth]{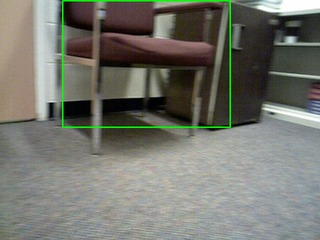}&   
		\includegraphics[width=0.22\linewidth]{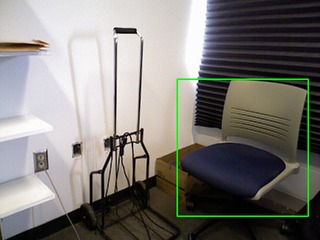}&
		\includegraphics[width=0.22\linewidth]{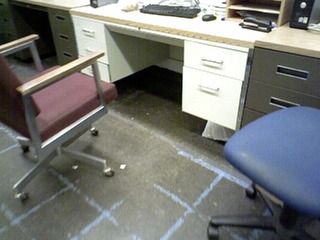}&
		\includegraphics[width=0.22\linewidth]{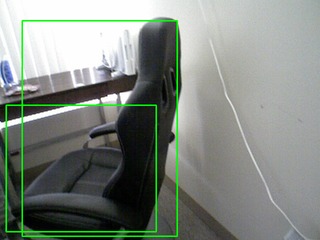}\\
		\includegraphics[width=0.22\linewidth]{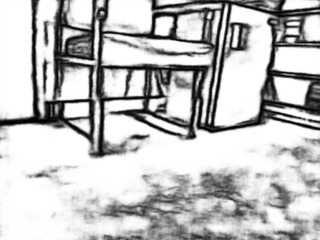}&   
		\includegraphics[width=0.22\linewidth]{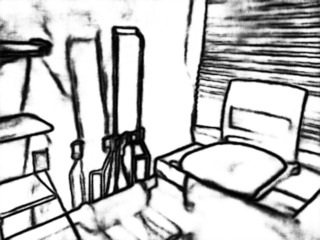}&
		\includegraphics[width=0.22\linewidth]{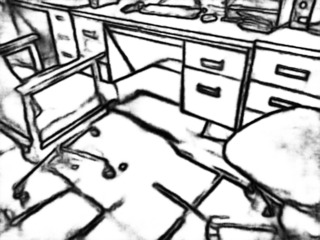}&
		\includegraphics[width=0.22\linewidth]{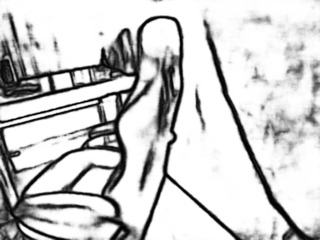}\\
		\includegraphics[width=0.22\linewidth]{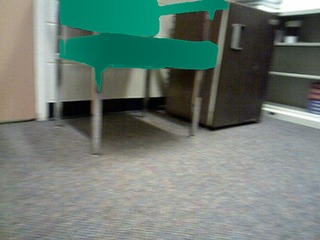}&   
		\includegraphics[width=0.22\linewidth]{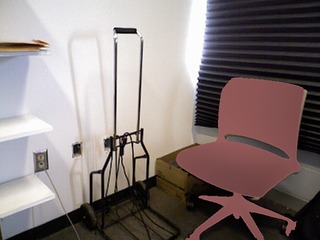}&
		\includegraphics[width=0.22\linewidth]{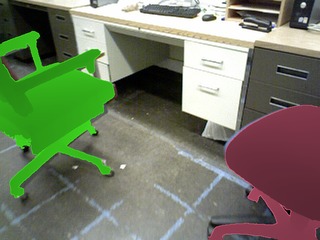}&
		\includegraphics[width=0.22\linewidth]{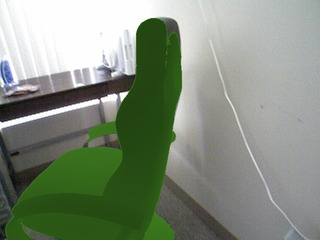}\\
		\includegraphics[width=0.22\linewidth]{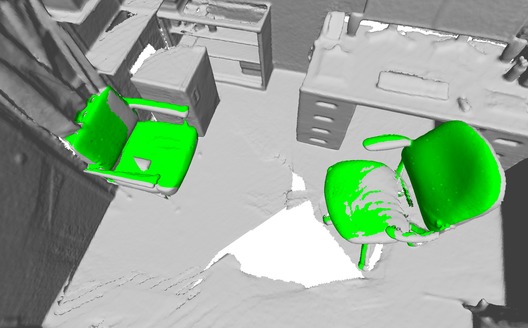}&   
		\includegraphics[width=0.22\linewidth]{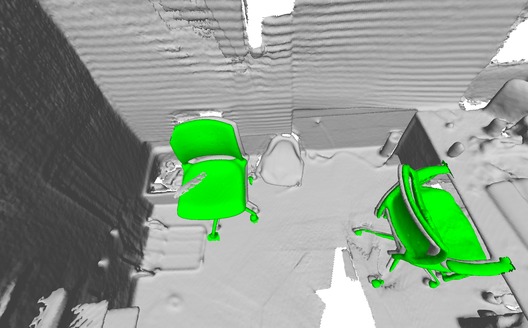}&   
		\includegraphics[width=0.22\linewidth]{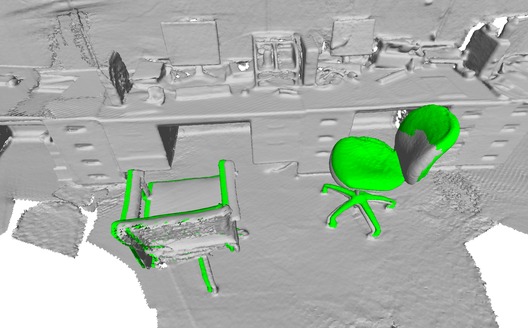}&
		\includegraphics[width=0.22\linewidth]{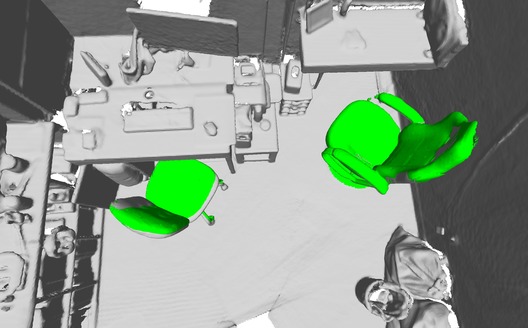}\\
		\scriptsize{025 (motion blur)} & \scriptsize{043 (distraction)} & \scriptsize{036 (missed detection)}  & \scriptsize{096 (duplicate)}
	\end{tabular}
\end{center}
	\vspace{-15pt}
	\caption{\captionfont {\it Qualitative results on SceneNN.} (best in color at $5 \times$) Each panel has the same meaning as Fig.~\ref{fig-visual}. Last row shows estimated shape \& pose (green) overlaid on ground truth mesh (gray).
    Partial projections due to broken models provided by SceneNN. 
	1st col: Moderate motion blur does not affect edge extraction. 
	2nd col: Background distraction does not affect shape \& pose inference thanks to the holistic and semantic knowledge injected into low-level edge features. 
	3rd col: Missed detections due to truncation resolved by memory. 
	4th col: Duplicate detection from Faster R-CNN eliminated by memory and inference in a consistent spatial frame.}
	\label{fig-scenenn-visual}
	\vspace{-20pt}
\end{figure}
\noindent{\bf Relation to the Prior Art}
Many efforts have been made to incorporate semantics into SLAM, and vice versa. Early attempts~\cite{castle2010combining,civera2011towards} rely on feature matching to register 3D objects to point clouds, which are sensitive to illumination and viewpoint changes, and most importantly, cannot handle texture-less objects. These issues are resolved by considering both semantic and geometric cues in our method (Fig.~\ref{fig-visual} and~\ref{fig-scenenn-visual}). In~\cite{kundu2014joint}, voxel-wise semantic labeling is achieved by fusing sparse reconstruction and pixel-wise semantic segmentation with a CRF model over voxel grids. The same scheme has been adopted by~\cite{hermans2014dense,vineet2015incremental,mccormac2017semanticfusion} which explore different sensors to get better reconstruction. Although these methods produce visually pleasing semantic labeling at the level of voxels, object-level semantic understanding is missing without additional steps to group together the potentially over-segmented voxels. Our method treats objects in the scene as first-class citizens and places objects in the scene directly and immediately without post-processing. The works that are closest to ours are RGB-D based SLAM++~\cite{slampp} and visual-inertial based~\cite{dong2017visual} and~\cite{bowman2017probabilistic}, where the former models objects as generic parallelepipeds and the latter focuses on the data association problem and only estimates translation of objects, while ours estimates precise object shape and 6DoF pose. 

This work is related to visual-inerital sensor fusion~\cite{mourikisr07} and vision-only monocular SLAM~\cite{klein2007parallel} in a broader sense. While classic SLAM outputs a descriptor-attached point cloud for localization, ours also populates objects in the scene to enable augmented reality (AR) and robotic tasks.

This work, by its nature, also relates to recent advances in object detection, either in two stages~\cite{girshick2015fast,ren2015faster,he2017maskrcnn}, which consist of a proposal generation and a regression/classification step, or in a single shot~\cite{liu2016ssd,redmon2016you}, where pre-defined anchors are used. Though single-shot methods are in general faster than two-stage methods, the clear separation of the architecture in the latter suits our hypothesis testing framework better (Fig.~\ref{fig-flowchart}). Image-based object detectors have encouraged numerous applications, however they are insufficient to fully describe the 3D attributes of objects. Efforts in making 2D detectors capable of 6DoF pose estimation include~\cite{xiang_wacv14,objectnet3d}, which are single image based and do not appreciate a globally consistent spatial reference frame, in which evidence can be accumulated over time as we did in our system.

The idea of using edge as a likelihood to estimate object pose dates back to the RAPiD algorithm~\cite{drummond2002real} followed by~\cite{klein2006full,choi20123d}. \cite{lepetit2005monocular} is a recent survey on model-based tracking, which is a special and simplistic case of our system: In model-based tracking, the 3D model being tracked is selected and its pose initialized manually while in our setting, such quantities are found by the algorithm. Another line of work~\cite{prisacariu2012pwp3d,tjaden2017real} on model-based tracking relies on level-set and appearance modeling, which we do not adopt because appearance is subject to illumination and viewpoint changes while edges are geometric and more robust.

\subsubsection*{Acknowledgment}
Research supported by ONR N00014-17-1-2072 and ARO W911NF-17-1-0304.
\clearpage
\bibliographystyle{splncs}

\end{document}